\documentclass{article}
\usepackage{xcolor}
\usepackage{graphicx}

\title{Bias Discovery in Machine Learning Models for Mental Health}

\author{Pablo Mosteiro $^{1}$, Jesse Kuiper $^{1}$, Judith Masthoff $^{1}$, \\
  Floortje Scheepers $^{2}$ and Marco Spruit $^{1,3,4}$}

\date{%
$^{1}$ \quad Department of Information and Computing Sciences, Utrecht University, Utrecht, the Netherlands; \{p.mosteiro,j.f.m.masthoff\}@uu.nl, jesse94kuiper@gmail.com \\
$^{2}$ \quad Afdeling Psychiatrie, University Medical Center Utrecht, Utrecht, the Netherlands; F.E.Scheepers-2@umcutrecht.nl \\
$^{3}$ \quad Department of Public Health and Primary Care, Leiden University Medical Center,
Leiden, the Netherlands; M.R.Spruit@lumc.nl \\
$^{4}$ \quad Leiden Institute of Advanced Computer Science, Leiden University, Leiden, the
Netherlands}

\begin{document}
\maketitle
\abstract{Fairness and bias are crucial concepts in Artificial Intelligence, yet they are relatively ignored in Machine Learning applications in clinical psychiatry. We compute fairness metrics and present bias mitigation strategies using a model trained on clinical mental health data. We collect structured data related to the admission, diagnosis and treatment of patients in the psychiatry department of the University Medical Center Utrecht. We train a machine learning model to predict future administrations of benzodiazepines on the basis of past data. We find that gender plays an unexpected role in the predictions; this constitutes bias. Using the AI Fairness 360 package, we implement reweighing and discrimination-aware regularization as bias mitigation strategies, and we explore their implications for model performance. This is the first application of bias exploration and mitigation to a machine learning model trained on real clinical psychiatry data.}




\section{Introduction}

For over ten years there has been increasing interest in the psychiatry domain for using Machine Learning~(ML) to aid psychiatrists and nurses~\cite{Pestian2010}. Recently, multiple approaches have been tested for Violence Risk Assessment (VRA)~\cite{Menger2019, Le2018, Suchting2018}, suicidal behaviour prediction~\cite{VanMens2020}, and the prediction of involuntary admissions~\cite{Kalidas2020}, among others.

Using ML for clinical psychiatry is appealing both as a time-saving instrument and as a way to provide insights to clinicians that might otherwise remain unexploited. Clinical ML models are usually trained on patient data, which includes some protected attributes, such as gender or ethnicity.
We desire for models to give equivalent outputs for equivalent patients that differ only in the value of a protected attribute~\cite{DelgadoRodriguez2004}. Yet, a systematic assessment of the fairness of ML models used for clinical psychiatry is lacking in the literature.

As a case study, we focus on the task of predicting future administrations of benzodiazepines.
Benzodiazepines are prescription drugs used in the treatment of, for example, anxiety and insomnia. Long-term use of benzodiazepines is associated with increased medical risks, such as cancer~\cite{Kim2017}. In addition, benzodiazepines in high doses are addictive, with complicated withdrawal~\cite{Quaglio2012}.
From a clinical perspective, gender should not play a role in the prescription of benzodiazepines~\cite{FederatieMedischSpecialisten2010, Vinkers2012}.
Yet biases in the prescription of benzodiazepines have been explored extensively in the literature; some protected attributes that contributed to bias were prescriber gender~\cite{Bjorner2003}, patient ethnicity~\cite{Peters2015, Cook2018}, and patient gender~\cite{Olfson2015}, as well as interaction effects between some of these protected attributes~\cite{McIntyre2020, Lui2021}. There is no conclusive concensus regarding these correlations, with some studies finding no correlations between socio-demographic factors and benzodiazepines prescriptions~\cite{Maric2017}.

We explore the effects of gender fairness bias on a model trained to predict the future administration of benzodiazepines to psychiatric patients based on past data, including past doses of benzodiazepines.
A possible use case of this model is to identify patients that are at risk of taking benzodiazepines for too long.
We hypothesize that our model is likely to unfairly use the patient's gender in making predictions. If that is the case, then mitigation strategies must be put in place to reduce this bias. We expect that there will be a cost to predictive performance.

Our research questions are:
\begin{enumerate}
\item For a model trained to predict future administrations of benzodiazepines based on past data, does gender unfairly influence the decisions of the model?
\item If gender does influence the decisions of said model, how much model performance is sacrificed when applying mitigation strategies to avoid the bias?
\end{enumerate}
To answer these questions, we employ a patient dataset from the University Medical Center (UMC) Utrecht and train a model to predict future administrations of benzodiazepines. We apply the bias discovery and mitigation toolbox AI Fairness 360~\cite{Bellamy2019}.
Whenever we find that gender bias is present in our model, we present an appropriate way to mitigate this bias.
\textcolor{black}{Our main contribution is a first implementation of a fairness evaluation and mitigation framework on real-world clinical data from the psychiatry domain. We present a way to mitigate a real and well known bias in benzodiazepine prescriptions, without loss of performance.}

\textcolor{black}{In Section~\ref{sec:2} we describe our materials and methods, including a review of previous work in the field. In Section~\ref{sec:3} we present our results, which we discuss in Section~\ref{sec:4}. We present our conclusions in Section~\ref{sec:5}.}
\section{Materials and Methods}
\label{sec:2}
\subsection{Related Work}
The study of bias in machine learning has garnered attention for several years~\cite{Baer2019}. 
\cite{Ellenberg1994} outlined the dangers of selection bias. Even when researchers attempt to be unbiased, problems might arise, such as bias from an earlier work trickling down into a new model~\cite{Barocas2016} or implicit bias from variables correlated with protected attributes~\cite{Dalessandro2017, Lang1993}.
\cite{Chouldechova2020} reviewed bias in machine learning, noting also that there is no industry standard for the definition of \emph{fairness}.
\cite{Dwork2011} evaluated bias in a machine learning model used for university admissions; they also point out the difference between \emph{individual} and \emph{group} fairness, as do~\cite{Zemel2013}. \cite{Joseph2016} and~\cite{Friedler2016} provided theoretical frameworks for the study of fairness.
Along the same lines, \cite{Saleiro2018} and \cite{Feldman2015} provided metrics for the evaluation of fairness. \cite{Kamiran2012} and \cite{Kamishima2012} recommend methods for mitigating bias.

As for particular applications, \cite{Scheuerman2020}, \cite{Xu2020} and~\cite{Yucer2020} studied race and gender bias in facial analysis systems. \cite{Liu2020} evaluated fairness in dialogue systems. While they did not actually evaluate ML models, \cite{Kizilcec2021} highlighted the importance of bias mitigation in AI for education.

In the medical domain, \cite{Genevieve2020} pointed out the importance of bias mitigation. Indeed, \cite{Tripathi2020} uncovered bias in post-operative complication predictions. \cite{Singh2021} found that disparities metrics change when transferring models across hospitals. Finally, \cite{Amir2021} explored the impact of random seeds on the fairness of classifiers using clinical data from MIMIC-III, and found that small sample sizes can also introduce bias.

No previous study on ML fairness or bias focuses on the psychiatry domain. This domain is interesting because bias seems to be present in the daily practice. We have already discussed in the introduction how bias is present in the prescription of benzodiazepines. There are also gender disparities in the prescription of zolpidem~\cite{Jasuja2019} and in the act of seeking psychological help~\cite{Nam2010}. \cite{Strakowski1996} also found racial disparities in clinical diagnoses with mania. Furthermore, psychiatry is a domain where a large amount of data is in the form of unstructured text, which is starting to be exploited for ML solutions~\cite{Rumshisky2016, Tang2021}. 
\textcolor{black}{Previous work has also focused on the explainability of text-based computational support systems in the psychiatry domain~\cite{kaczmarek}.}
It will be crucial as these text-based models begin to be applied in the clinical practice to ensure that they too are unbiased towards protected attributes.

\subsection{Data}
\label{sec:22}
We employ de-identified patient data from Electronic Health Records (EHRs) from the psychiatry department at the UMC Utrecht. Patients in the dataset were admitted to the psychiatry department between June 2011 and May 2021. The five database tables included are: admissions, patient information, medication administered, diagnoses, and violence incidents. Table~\ref{tab:1} shows the variables present in each of the tables.
\begin{table}
\centering
\caption{Datasets retrieved from the psychiatry department of the UMC Utrecht, with the variables present in each dataset that are used for this study. Psychiatry is divided into four \emph{nursing wards}. For the ``medication'' dataset, the ``Administered'' and ``Not administered'' variables contain in principle the same information; however, sometimes only one of them is filled.}
\label{tab:1}
\begin{tabular}{|l|l|l|} 
  \hline
  Dataset & Variable & Type \\
  \hline
  \hline
  Admissions & Admission ID & Identifier \\
  & Patient ID & Identifier \\
  & Nursing ward ID & Identifier \\
  & Admission date & Date \\
  & Discharge date & Date \\
  & Admission time & Time \\
  & Discharge time & Time \\
  & Emergency & Boolean \\
  & First admission & Boolean \\
  & Gender & Man / Woman \\
  & Age at admission & Integer \\
  & Admission status & Ongoing / Discharged \\
  & Duration in days & Integer \\
  \hline
    Medication & Patient ID & Identifier \\
  & Prescription ID & Identifier \\
  & ATC code (medication ID) & String \\
  & Medication name & String \\
  & Dose & Float \\
  & Unit (for dose) & String \\
  & Administration date & Date \\
  & Administration time & Time \\
  & Administered & Boolean \\
  & Dose used & Float \\
  & Original dose & Float \\
  & IsContinuationAfterSuspension & Boolean \\
  & Not administered & Boolean \\
  \hline
             Diagnoses & Patient ID & Identifier \\
  & Diagnosis number & Identifier \\
  & Start date & Date \\
  & End date & Date \\
  & Main diagnosis group & Categorical \\
  & Level of care demand & Numeric \\
  & Multiple problem & Boolean \\
  & Personality disorder & Boolean \\
  & Admission & Boolean \\
  & Diagnosis date & Date \\
  \hline
    Aggression & Patient ID & Identifier \\
  & Date of incident & Date \\
  & Start time & Time \\
  \hline
      Patient & Patient ID & Identifier \\
    & Age at start of dossier & Integer\\
  \hline
\end{tabular}
\end{table}

We construct a dataset where each data point is 14 days after the admission of a patient. We select only completed admissions (admission status = ``discharged'') that lasted at least 14 days (duration in days~$\geq$~14). 3192 admissions (\textit{i.e.}, data points) are included in our dataset. These are coupled with data from the other four tables mentioned above. The nursing ward ID is converted to four binary variables; some rows do not belong to any nursing ward ID (because, for example, the patient was admitted outside of psychiatry and then transferred to psychiatry); these rows have zeros for all four nursing ward ID columns.

For diagnoses, the diagnosis date was not always present in the dataset. In that case, we used the end date of the treatment trajectory. If that was also not present, we used the start date of the treatment trajectory. One of the entries in the administered medication table had no date of administration; this entry was removed. We only consider administered medication (administered = {\tt True}). Doses of various tranquilizers are converted to an equivalent dose of diazepam, according to Table~\ref{tab:2}~\cite{Nhg2014}.\footnote{
\textcolor{black}{This is the normal procedure when investigating benzodiazepine use. All benzodiazepines have the same working mechanism. The only differences are the half-life and the peak time. So when studying benzodiazepines it is allowed to make an equivalent dose of one specific benzodiazepine.}
}
\begin{table}
\centering
\caption{List of tranquilizers considered in this study, along with the multipliers used for scaling the doses of those tranquilizers to a diazepam-equivalent. The last column is the inverse of the center column.}
\label{tab:2}
\begin{tabular}{|l|c|c|} 
  \hline
  Tranquilizer & Multiplier & mg / (mg diazepam) \\
  \hline
  \hline
Diazepam  &  1.0  &  1.00  \\
Alprazolam  &  10.0  &  0.10  \\
Bromazepam  &  1.0  &  1.00  \\
Brotizolam  &  40.0  &  0.03  \\
Chlordiazepoxide  &  0.5  &  2.00  \\
Clobazam  &  0.5  &  2.00  \\
Clorazepate potassium  &  0.75  &  1.33  \\
Flunitrazepam          &  0.1   &  10    \\
Flurazepam  &  0.33  &  3.03  \\
Lorazepam  &  5.0  &  0.20  \\
Lormetazepam  &  10.0  &  0.10  \\
Midazolam  &  1.33  &  0.10  \\
Nitrazepam  &  1.0  &  1.00  \\
Oxazepam  &  0.33  &  3.03  \\
Temazepam  &  1.0  &  1.00  \\
Zolpidem  &  1.0  &  1.00  \\
Zopiclone  &  1.33  &  0.75  \\
\hline
\end{tabular}
\end{table}

For each admission, we obtain the age of the patient at the start of the dossier from the patient table. The gender is reported in the admissions table; only the gender assigned at birth is included in this dataset. We count the number of violence incidents before admission and the number of violence incidents during the first 14 days of admission. The main diagnosis groups were converted to binary values, where 1 means that this diagnosis was present for that admission, and that it took place during the first 14 days of admission. Other binary variables derived from the diagnoses table are ``Multiple problem'' and ``Personality disorder''. For all diagnoses present for a given admission, we computed the maximum and minimum ``levels of care demand'', and saved them as two new variables. Matching the administered medication to the admissions by patient ID and date, we compute the total amount of diazepam-equivalent benzodiazepines administered in the first 14 days of admission, and the total administered in the remainder of the admission. The former is one of the predictor variables.
The target variable is binary, \textit{i.e.}, whether benzodiazepines are administered during the remainder of the admission or not.

The final dataset consists of 3192 admissions. 1724 admissions correspond to men, while 1468 are women. 2035 admissions have some benzodiazepines administered during the first 14 days of admission, while 1980 admissions have some benzodiazepines administered during the remainder of the admission. Table~\ref{tab:3} shows the final list of variables included in the dataset.
\begin{table}
\centering
\caption{List of variables in the final dataset}
\label{tab:3}
\begin{tabular}{|l|c|} 
  \hline
  Variable & Type \\
  \hline
  \hline
Patient ID  &  Numeric  \\
Emergency  &  Binary  \\
First admission  &  Binary  \\
Gender  &  Binary  \\
Age at admission  &  Numeric  \\
Duration in days  &  Numeric  \\
Age at start of dossier  &  Numeric  \\
Incidents during admission  &  Numeric  \\
Incidents before admission  &  Numeric  \\
Multiple problem  &  Binary  \\
Personality disorder  &  Binary  \\
Minimum level of care demand  &  Numeric  \\
Maximum level of care demand  &  Numeric  \\
Past diazepam-equivalent dose  &  Numeric  \\
Future diazepam-equivalent dose  &  Numeric  \\
Nursing ward: Clinical Affective \& Psychotic Disorders  &  Binary  \\
Nursing ward: Clinical Acute \& Intensive Care  &  Binary  \\
Nursing ward: Clinical Acute \& Intensive Care Youth  &  Binary  \\
Nursing ward: Clinical Diagnosis \& Early Psychosis  &  Binary  \\
Diagnosis: Attention Deficit Disorder  &  Binary  \\
Diagnosis: Other issues that may be a cause for concern  &  Binary  \\
Diagnosis: Anxiety disorders  &  Binary  \\
Diagnosis: Autism spectrum disorder  &  Binary  \\
Diagnosis: Bipolar Disorders  &  Binary  \\
Diagnosis: Cognitive disorders  &  Binary  \\
Diagnosis: Depressive Disorders  &  Binary  \\
Diagnosis: Dissociative Disorders  &  Binary  \\
Diagnosis: Behavioral disorders  &  Binary  \\
Diagnosis: Substance-Related and Addiction Disorders  &  Binary  \\
Diagnosis: Obsessive Compulsive and Related Disorders  &  Binary  \\
Diagnosis: Other mental disorders  &  Binary  \\
Diagnosis: Overige stoornissen op zuigelingen of kinderleeftijd  &  Binary  \\
Diagnosis: Personality Disorders  &  Binary  \\
Diagnosis: Psychiatric disorders due to a general medical condition  &  Binary  \\
Diagnosis: Schizophrenia and other psychotic disorders  &  Binary  \\
Diagnosis: Somatic Symptom Disorder and Related Disorders  &  Binary  \\
Diagnosis: Trauma- and stressor-related disorders  &  Binary  \\
Diagnosis: Nutrition and Eating Disorders  &  Binary  \\
\hline
\end{tabular}
\end{table}

\subsection{Evaluation Metrics}
\label{sec:23}
The performance of the model is to be evaluated by the use of the balanced accuracy~\footnote{\textcolor{black}{As seen in Section~\ref{sec:22}, the distribution of data points across classes is almost balanced. With that in mind, we could have used accuracy instead of balanced accuracy. However, we had decided on an evaluation procedure before looking at the data, based on previous experience in the field. We find no reason to believe that our choice should affect the results significantly.}} (average of true positive rate and true negative rate) and the F1 score. As for quantifying bias, we will use \textcolor{black}{four} metrics:
\begin{itemize}
\item \emph{Statistical Parity Difference} is discussed in \cite{Dwork2011}, as the difference between the correctly classified instances for the privileged and the unprivileged group. If the statistical parity difference is 0, privileged and unprivileged groups receive the same percentage of positive classifications. Statistical parity is an indicator for representation and therefore a group fairness metric. If the value is negative, the privileged group has an advantage.
\item \emph{Disparate Impact} is computed as the ratio of rate of favourable outcome for the unprivileged group to that of the privileged group~\cite{Feldman2015}. This value should be as close to 1 for a fair result; lower than 1 implies a benefit for the privileged group.
\item \emph{Equal Opportunity Difference} is the difference between the true positive rates between the unprivileged group and the privileged group. It evaluates the ability of the model to classify the unprivileged group compared to the privileged group. The value should be as close to 0 for a fair result. If the value is negative, the privileged group has an advantage.
\item \emph{Average Odds Difference} is the difference between false positives rates and true positive rates between the unprivileged group and privileged group. It provides insights into a possible positive biases towards a group. This value should be as close to 0 for a fair result. If the value is negative, the privileged group has an advantage.
\end{itemize}

\subsection{Machine Learning Methods}
\label{sec:25}
We use AI Fairness 360, a package for the discovery and mitigation of bias in machine learning models. The protected attribute in our dataset is gender, while the favourable class is ``man''.
We employ two classification algorithms implemented in ScikitLearn~\cite{Pedregosa2011}: logistic regression and random forest~\footnote{\textcolor{black}{We consider these models because they are simple, widely available and widely used within and beyond the clinical field.}}. For logistic regression, we use the ``liblinear'' solver. For the random forest classifier, we use 500 estimators, with min\_samples\_leaf equal to 25.

\begin{figure}[!ht]
\begin{center}
  \includegraphics[width=10cm]{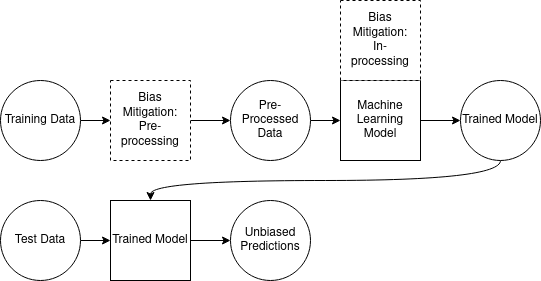}
\end{center}
\caption{\textcolor{black}{Workflow of data, machine learning models and bias mitigation techniques used in this research.}}
\label{fig:workflow}
\end{figure}

There are three types of bias mitigation techniques: \emph{pre-processing}, \emph{in-processing}, and \emph{post-processing}~\cite{Dalessandro2017}. Pre-processing techniques mitigate bias by removing the underlying discrimination from the dataset. In-processing techniques are modifications to the machine learning algorithms to mitigate bias during model training. Post-processing techniques seek to mitigate bias by equalizing the odds post-training.  
We use two methods for bias mitigation. As a \emph{pre-processing} method, we use the reweighing technique of \cite{Kamiran2012}, and re-train our classifiers on the reweighed dataset. As an \emph{in-processing} method, we add a discrimination-aware regularization term to the learning objective of the logistic regression model. This is called a \emph{prejudice remover}. We set the fairness penalty parameter {\tt eta} to 25\textcolor{black}{, which is high enough that prejudice will be removed aggressively, while not too high such that accuracy will be significantly compromised~\cite{Kamishima2012}}.
Both of these techniques are seamlessly implemented in AI Fairness 360.
\textcolor{black}{To apply \emph{post-processing} techniques in practice, one needs a training set and a test set; once the model is trained, the test set is used to determine how outputs should be modified in order to limit bias. However, in clinical applications datasets tend to be small, so we envision a realistic scenario in which the entire dataset is used for development, making the use of post-processing methods impossible. For this reason, we do not study these methods further.}
\textcolor{black}{The workflow of data, models and bias mitigation techniques is shown on Figure~\ref{fig:workflow}.}

\textcolor{black}{To estimate the uncertainty due to the choice of training data, we use 5-fold cross-validation, with patient IDs as group identifiers to avoid using the same sample for development and testing. Within each fold, we again split the development set into 62.5\% training, 37.5\% validation, once again with patient IDs as group identifiers to avoid using the same sample for training and validation. We train the model on the training set, and use the validation set to compute the optimal classification threshold, which is the threshold that maximizes the balanced accuracy on the validation set. We then re-train the model on the entire development set, and compute the performance and fairness metrics on the test set. Finally, we compute the mean and standard deviation of all metrics across the 5 folds.}

The code used to generate the dataset and train the machine learning models is provided as a GitHub repository~\footnote{https://github.com/PabloMosUU/FairnessForPsychiatry}.

\section{Results}

\label{sec:3}
\begin{figure}[!ht]
\begin{center}
  \includegraphics[width=10cm]{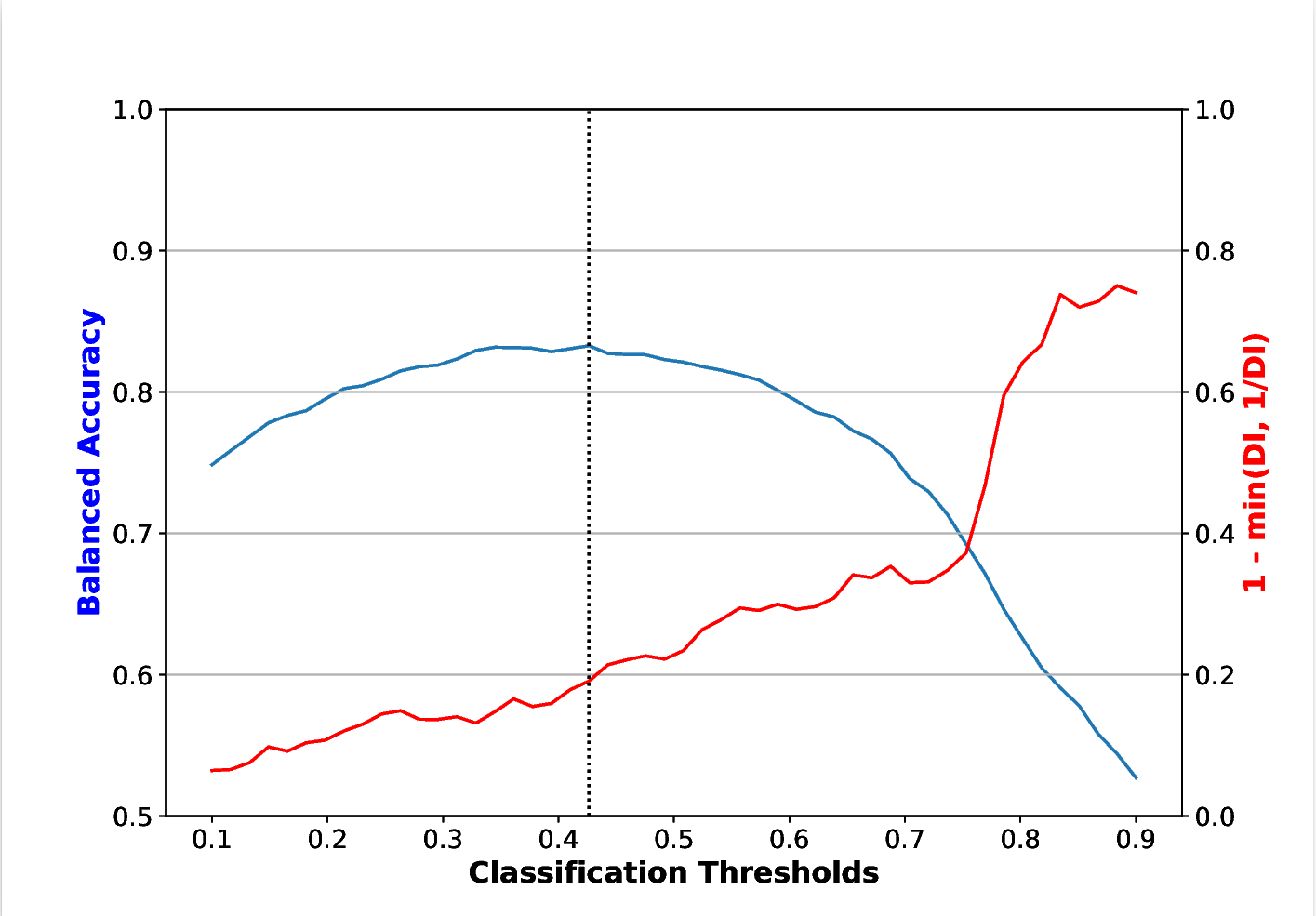}
\end{center}
\caption{Balanced accuracy and disparate impact error versus classification threshold for a logistic regression classifier with no bias mitigation. The dotted vertical line is the threshold that maximizes balanced accuracy. The plot shown corresponds to one of the folds of cross-validation. \textcolor{black}{Disparate impact error, equal to 1 - {\tt min}(DI, 1/DI), where DI is the disparate impact, is the difference between disparate impact and its ideal value of 1.}}
\label{fig:1}
\end{figure}

\begin{figure}[!ht]
\begin{center}
  \includegraphics[width=10cm]{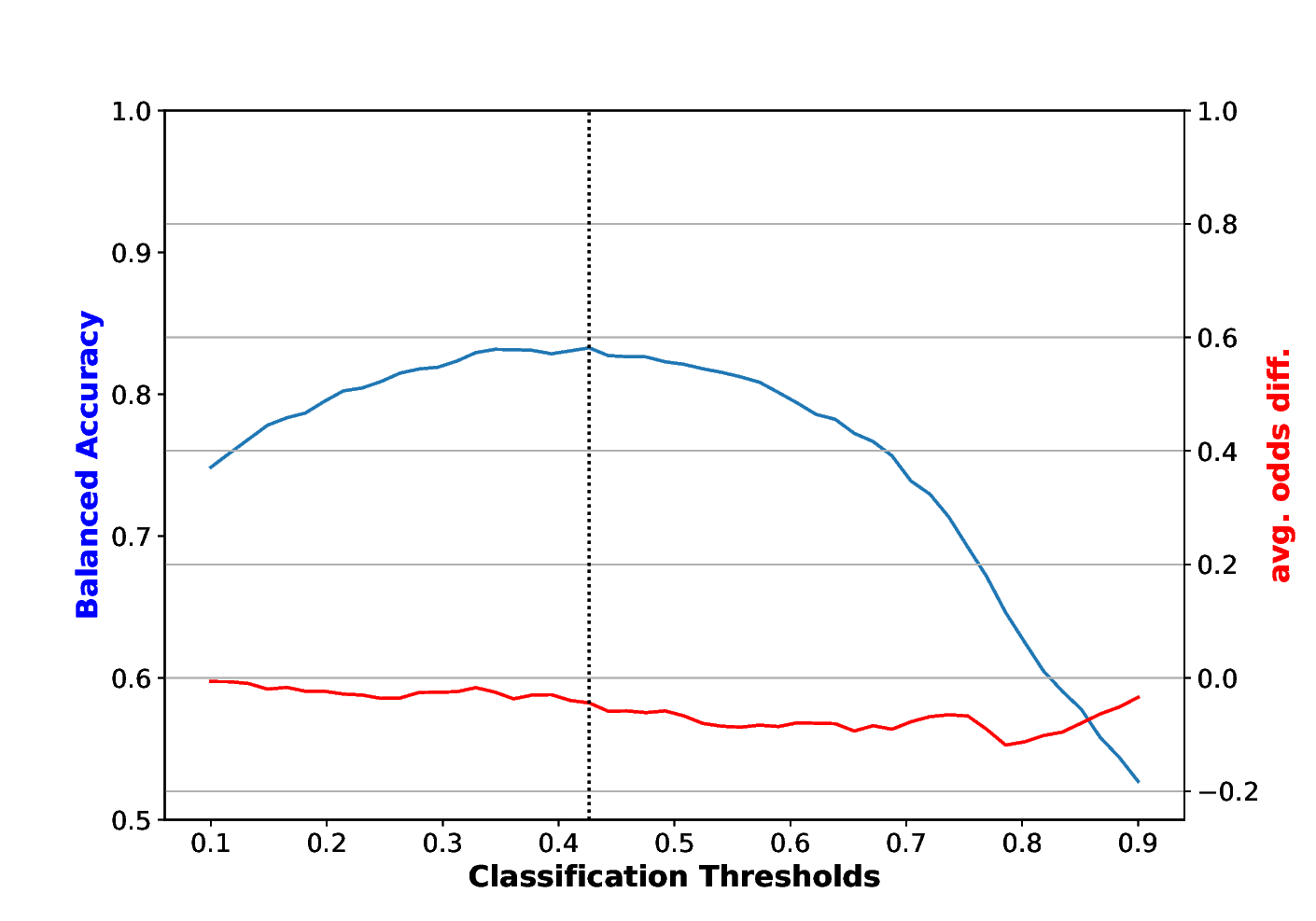}
\end{center}
\caption{Balanced accuracy and average odds difference versus classification threshold for a logistic regression classifier with no bias mitigation. The dotted vertical line is the threshold that maximizes balanced accuracy. The plot shown corresponds to one of the folds of cross-validation.}
\label{fig:2}
\end{figure}

\begin{figure}[!ht]
\begin{center}
  \includegraphics[width=10cm]{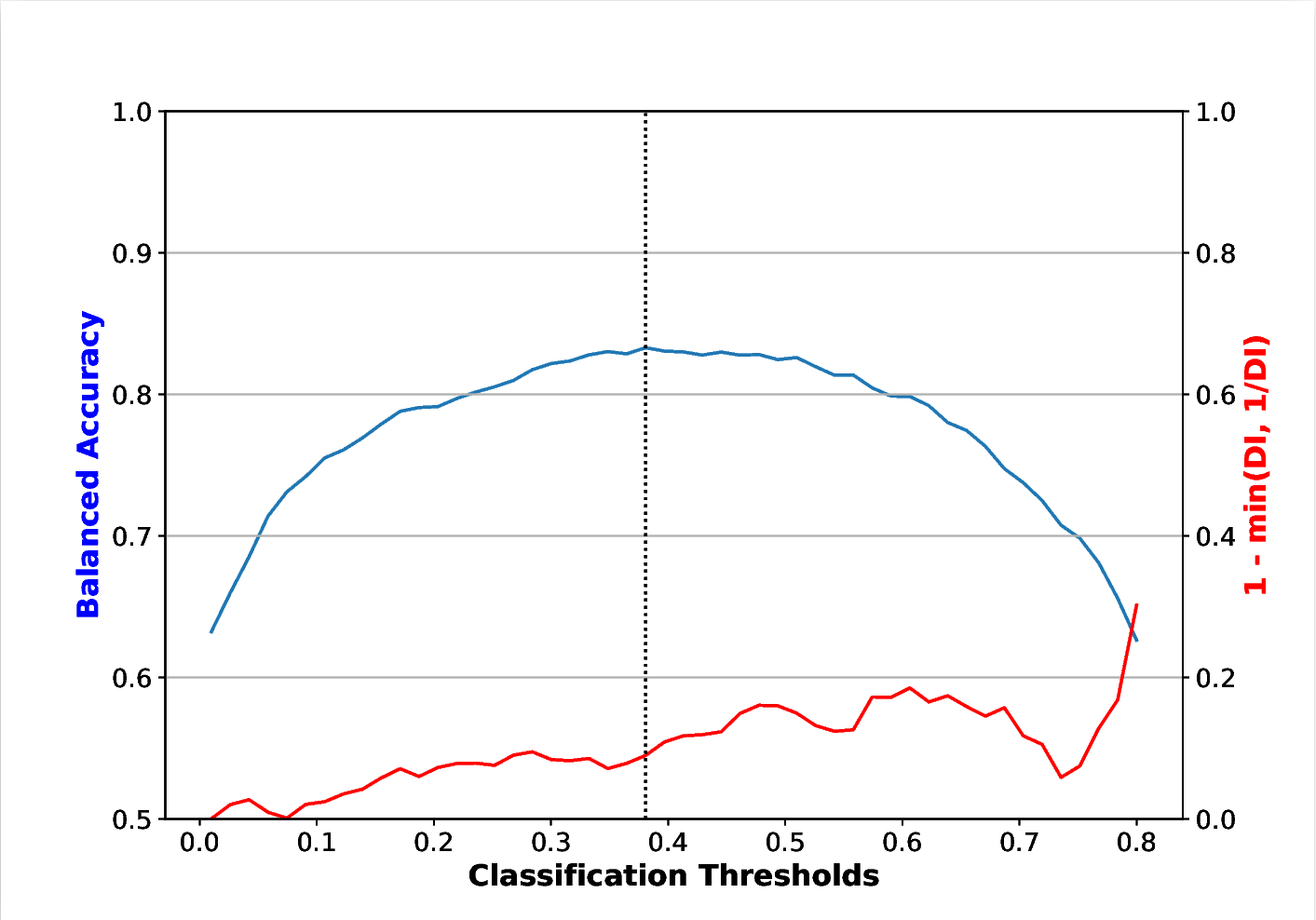}
\end{center}
\caption{Balanced accuracy and disparate impact error versus classification threshold for a logistic regression classifier with reweighing. The dotted vertical line is the threshold that maximizes balanced accuracy. The plot shown corresponds to one of the folds of cross-validation. \textcolor{black}{Disparate impact error, equal to 1 - {\tt min}(DI, 1/DI), where DI is the disparate impact, is the difference between disparate impact and its ideal value of 1.}}
\label{fig:3}
\end{figure}

\begin{figure}[!ht]
\begin{center}
  \includegraphics[width=10cm]{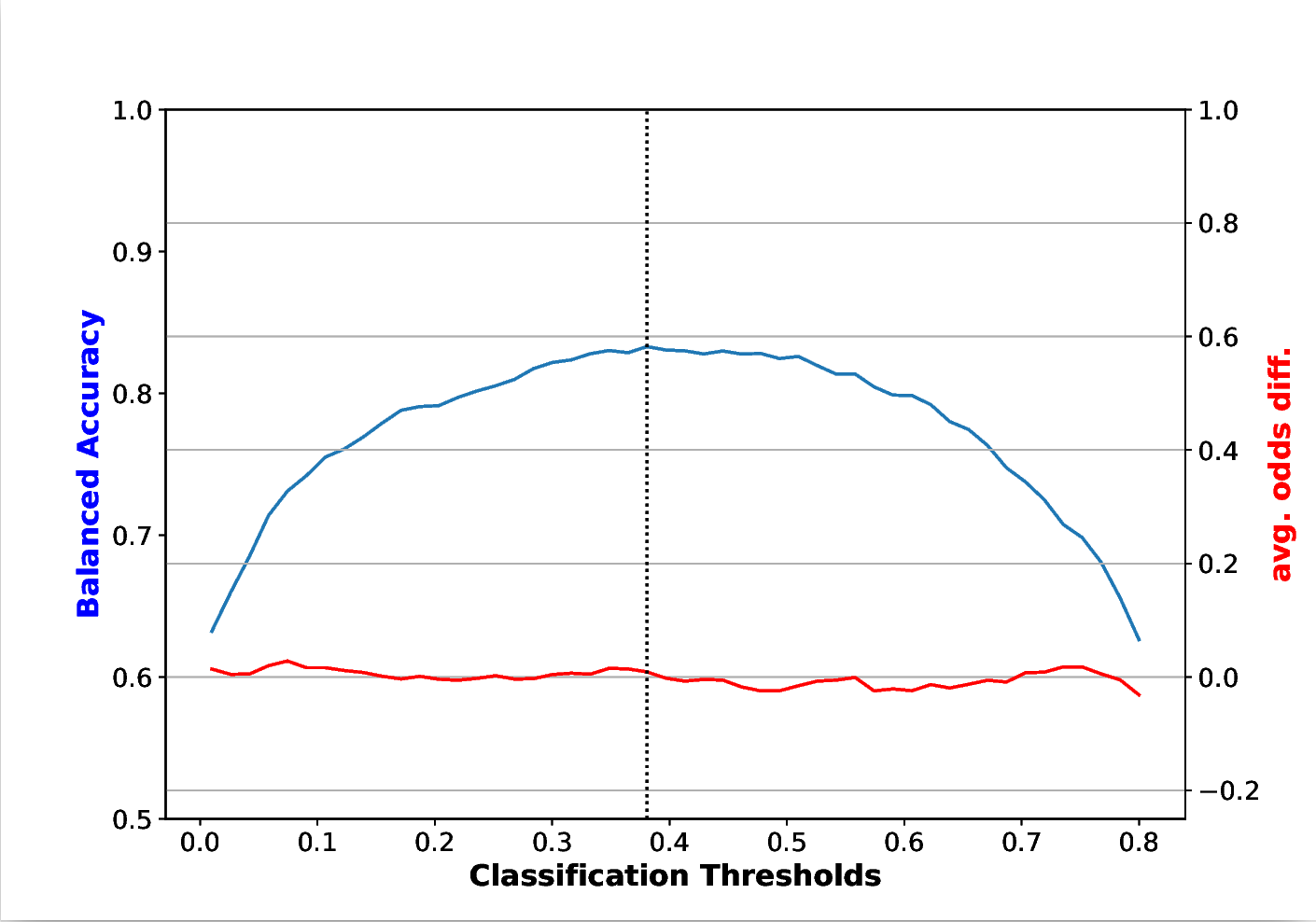}
\end{center}
\caption{Balanced accuracy and average odds difference versus classification threshold for a logistic regression classifier with reweighing. The dotted vertical line is the threshold that maximizes balanced accuracy. The plot shown corresponds to one of the folds of cross-validation.}
\label{fig:4}
\end{figure}

\begin{figure}[!ht]
\begin{center}
  \includegraphics[width=10cm]{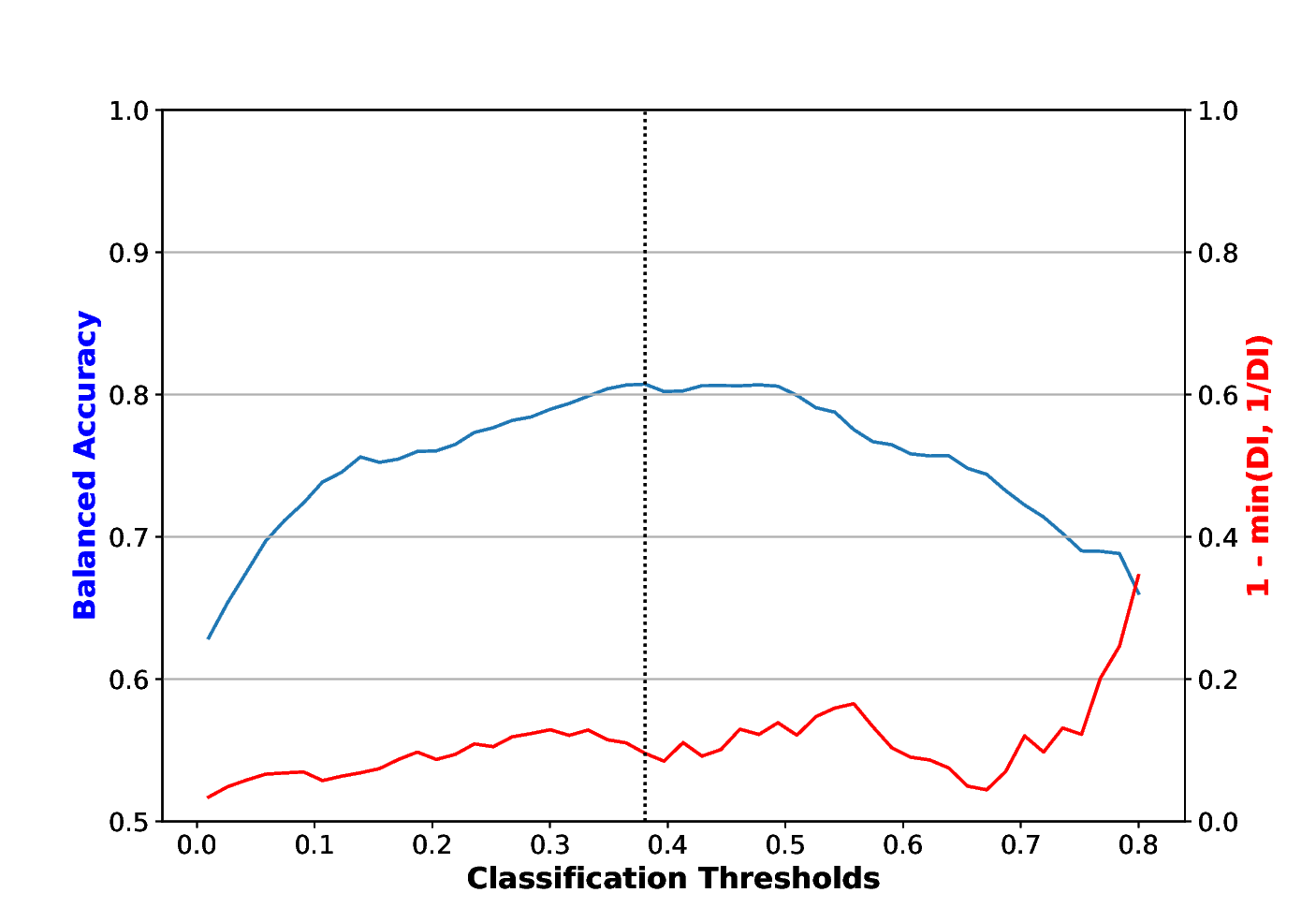}
\end{center}
\caption{Balanced accuracy and disparate impact error versus classification threshold for a logistic regression classifier with prejudice remover. The dotted vertical line is the threshold that maximizes balanced accuracy. The plot shown corresponds to one of the folds of cross-validation. \textcolor{black}{Disparate impact error, equal to 1 - {\tt min}(DI, 1/DI), where DI is the disparate impact, is the difference between disparate impact and its ideal value of 1.}}
\label{fig:5}
\end{figure}

\begin{figure}[!ht]
\begin{center}
  \includegraphics[width=10cm]{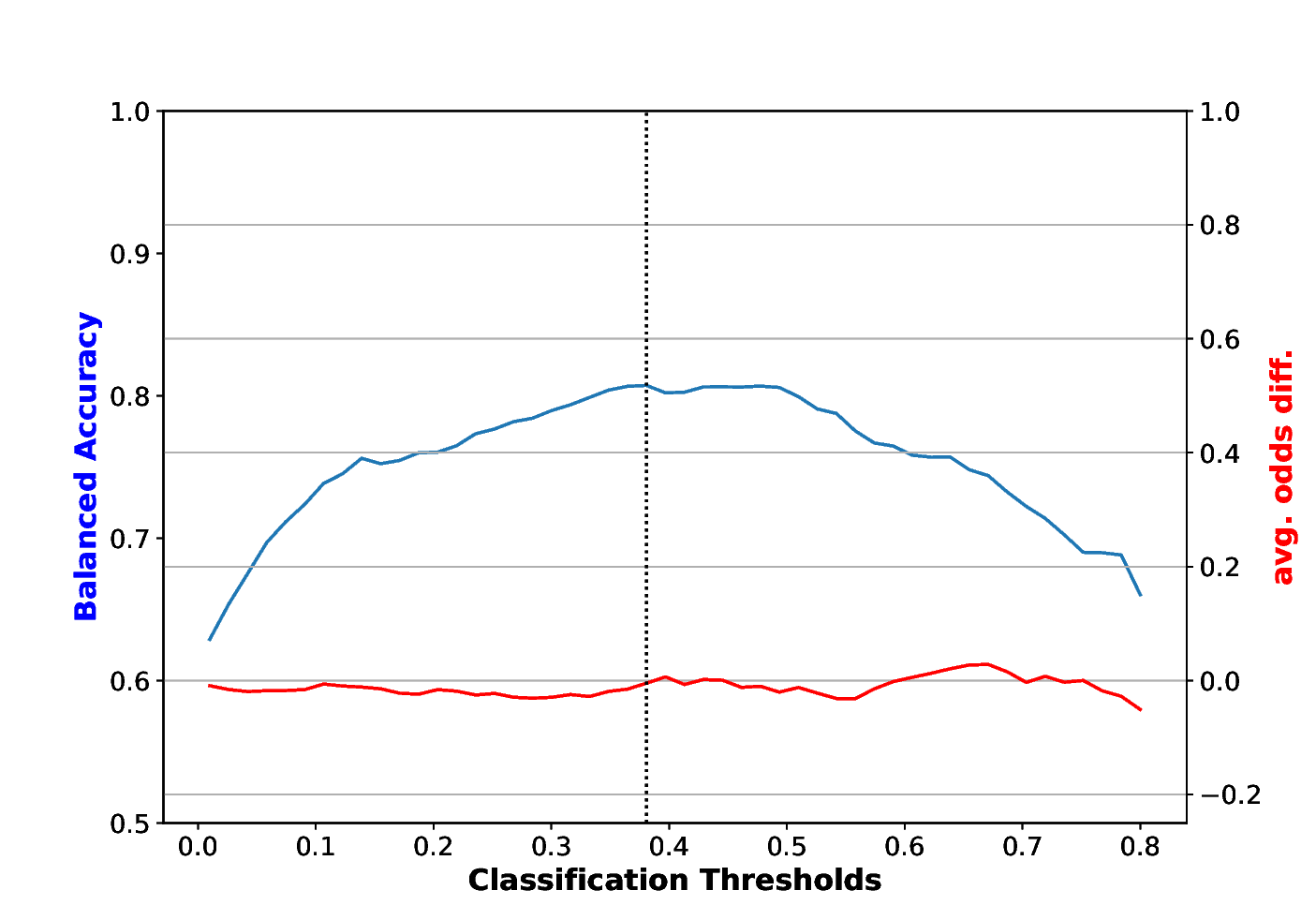}
\end{center}
\caption{Balanced accuracy and average odds difference versus classification threshold for a logistic regression classifier with prejudice remover. The dotted vertical line is the threshold that maximizes balanced accuracy. The plot shown corresponds to one of the folds of cross-validation.}
\label{fig:6}
\end{figure}

\textcolor{black}{Each of our classifiers outputs a continuous prediction for each test data point. We convert these to binary classifications by comparing with a classification threshold. Figures~\ref{fig:1} thru~\ref{fig:6} show the trade-off between balanced accuracy and fairness metrics as a function of the classification threshold.}
Figures~\ref{fig:1} and~\ref{fig:2} show how the disparate impact error and average odds difference vary together with the balanced accuracy as a function of the classification threshold of a logistic regression model with no bias mitigation\textcolor{black}{, for one of the folds of cross-validation}. The corresponding plots for the random forest classifier show the same trends.
The \textcolor{black}{performance and fairness metrics after cross-validation} are shown in Table\textcolor{black}{s}~\ref{tab:4}\textcolor{black}{ and}~\ref{tab:5}\textcolor{black}{, respectively}. Since we observe bias (see Section~\ref{sec:4} for further discussion), we implement the mitigation strategies detailed in Section~\ref{sec:25}. Figures~\ref{fig:3} and~\ref{fig:4} show the validation plots for a logistic regression classifier with reweighing \textcolor{black}{for one of the folds of cross-validation}; the plots for the random forest classifier show similar trends. Figures~\ref{fig:5} and~\ref{fig:6} show the validation plots for a logistic regression classifier with prejudice remover.
\begin{table}
  \centering
  \caption{\textcolor{black}{Classification metrics for logistic regression (LR) and random forest (RF) classifiers including bias mitigation strategies reweighing (RW) and prejudice remover (PR). The classification metrics are balanced accuracy (Acc$_{\rm bal}$) and F1 score. The errors shown are standard deviations.}}
  \label{tab:4}
\begin{tabular}{|l|l|c|c|}
\hline
\multicolumn{2}{|c|}{Model} & \multicolumn{2}{c|}{Performance} \\
\hline
Clf. & Mit. & Acc$_{\rm bal}$ & F1 \\
\hline
\hline
LR &  & 0.834 +/- 0.015 & 0.843 +/- 0.014 \\
RF &  & 0.843 +/- 0.018 & 0.835 +/- 0.020 \\
\hline
LR & RW & 0.830 +/- 0.014 & 0.839 +/- 0.011 \\
RF & RW & 0.847 +/- 0.019 & 0.840 +/- 0.020 \\
\hline
LR & PR & 0.793 +/- 0.020 & 0.802 +/- 0.029 \\
\hline
\end{tabular}
\end{table}
\section{Discussion}

\label{sec:4}

\subsection{Analysis of Results}
\label{sec:41}
As reported in Table~\ref{tab:5}, all fairness metrics show results favourable to the privileged group \textcolor{black}{(see Section~\ref{sec:23} for a discussion of the fairness metrics we use)}. Reweighing improved the fairness metrics for both classifiers. The Prejudice Remover also improved the fairness metrics, albeit at a cost in performance. There was no big difference in performance between the logistic regression and random forest classifiers. If fairness is crucial, the logistic regression classifier gives more options in terms of the mitigation strategies. The better mitigation strategy is the one closest to the data, for it requires less tinkering with the model, which can lead to worse explainability.
\begin{table}
  \centering
  \caption{\textcolor{black}{Fairness metrics for logistic regression (LR) and random forest (RF) classifiers including bias mitigation strategies reweighing (RW) and prejudice remover (PR). The fairness metrics are disparate impact (DI), average odds difference (AOD), statistical parity difference (SPD) and equal opportunity difference (EOD). The errors shown are standard deviations.}}
  \label{tab:5}
\begin{tabular}{|l|l|c|c|c|c|}
\hline
\multicolumn{2}{|c|}{Model} & \multicolumn{4}{c|}{Fairness} \\
\hline
Clf. & Mit. & DI & AOD & SPD & EOD \\
\hline
\hline
LR &  & 0.793 +/- 0.074 & -0.046 +/- 0.021 & -0.110 +/- 0.038 & -0.038 +/- 0.028 \\
RF &  & 0.796 +/- 0.071 & -0.018 +/- 0.017 & -0.083 +/- 0.031 & -0.013 +/- 0.035 \\
\hline
LR & RW & 0.869 +/- 0.066 & -0.003 +/- 0.013 & -0.066 +/- 0.035 & 0.004 +/- 0.034 \\
RF & RW & 0.830 +/- 0.077 & -0.004 +/- 0.023 & -0.070 +/- 0.034 & 0.001 +/- 0.043 \\
\hline
LR & PR & 0.886 +/- 0.056 & -0.008 +/- 0.003 & -0.060 +/- 0.034 & -0.020 +/- 0.045 \\
\hline
\end{tabular}
\end{table}

\textcolor{black}{In addition, we computed, for each fold of cross-validation, the difference for each performance and fairness metric between a model with a bias mitigator and the corresponding model without bias mitigation.
  We then take the mean and standard deviation of those differences, and report the results for performance and fairness metrics on Tables~\ref{tab:6} and~\ref{tab:7}, respectively.
  We can see that differences in performance for reweighting are mostly small, while the gains in fairness metrics are statistically significant at 95\% confidence level. Meanwhile, the prejudice remover incurs a greater cost in performance, with no apparent greater improvement to the fairness metrics.}
\begin{table}
  \centering
  \caption{\textcolor{black}{Classification metric differences of models with bias mitigators reweighing (RW) and prejudice remover (PR) compared to a baseline without bias mitigation, for logistic regression (LR) and random forest (RF) classifiers. The classification metrics are balanced accuracy (Acc$_{\rm bal}$) and F1 score. The errors shown are standard deviations. Differences significant at 95\% confidence level are shown in \textbf{bold}.}}
  \label{tab:6}
\begin{tabular}{|l|l|c|c|}
\hline
\multicolumn{2}{|c|}{Model} & \multicolumn{2}{c|}{Performance} \\
\hline
Clf. & Mit. & $\Delta$Acc$_{\rm bal}$ & $\Delta$F1 \\
\hline
\hline
LR & PR & \textbf{-0.040 $\pm$ 0.013} & -0.041 $\pm$ 0.025 \\
LR & RW & -0.003 $\pm$ 0.013 & -0.005 $\pm$ 0.013 \\
\hline
RF & RW & 0.003 $\pm$ 0.002 & \textbf{0.005 $\pm$ 0.001} \\
\hline
\end{tabular}
\end{table}
\begin{table}
  \centering
  \caption{\textcolor{black}{Fairness metric differences of models with bias mitigators reweighing (RW) and prejudice remover (PR) compared to a baseline without bias mitigation, for logistic regression (LR) and random forest (RF) classifiers. The fairness metrics are disparate impact (DI), average odds difference (AOD), statistical parity difference (SPD) and equal opportunity difference (EOD). The errors shown are standard deviations. Differences significant at 95\% confidence level are shown in \textbf{bold}.}}
  \label{tab:7}
\begin{tabular}{|l|l|c|c|c|c|}
\hline
\multicolumn{2}{|c|}{Model} & \multicolumn{4}{c|}{Fairness} \\
\hline
Clf. & Mit. & $\Delta$DI & $\Delta$AOD & $\Delta$SPD & $\Delta$EOD \\
\hline
\hline
LR & PR & \textbf{0.092 $\pm$ 0.036} & 0.038 $\pm$ 0.021 & \textbf{0.050 $\pm$ 0.019} & 0.018 $\pm$ 0.042 \\
LR & RW & \textbf{0.075 $\pm$ 0.021} & \textbf{0.043 $\pm$ 0.017} & \textbf{0.043 $\pm$ 0.014} & 0.042 $\pm$ 0.034 \\
\hline
RF & RW & \textbf{0.034 $\pm$ 0.013} & \textbf{0.014 $\pm$ 0.006} & \textbf{0.013 $\pm$ 0.006} & 0.014 $\pm$ 0.011 \\
\hline
\end{tabular}
\end{table}

\subsection{Limitations}

Some diagnoses did not have a diagnosis date filled out in the raw dataset.
In those cases, we used the treatment end date.
Some data points did not have a value for that variable either, and in those cases we used the treatment start date. This leads to an inconsistent definition of the diagnosis date, and hence to inconsistencies in the variables related to diagnoses during the first 14 days of admission. However, we re-did the analysis with only the diagnoses for which diagnosis dates were present in the raw data, and the results followed the same trends.

On a similar note, we removed a few medication administrations that did not have an administering date. A better solution would have been to remove all data corresponding to those patients, albeit at the cost of having fewer data points. We re-did the analysis in that configuration, and obtained similar results.

Finally, this work considered only diagnoses that took place within the first 14 days of admission. It might have been interesting to also consider diagnoses that took place \emph{before} admission. We leave this option for future work.

\subsection{Future Work}

The present work considered benzodiazepine prescriptions administered during the remainder of each patient's admission. To make the prediction task fairer for the computer, we could consider predicting benzodiazepines administered during a specific time window, for example, days 15 thru 28 of an admission.

Previous work noted a possible bias between the gender of the \emph{prescriber} and the prescriptions of benzodiazepines~\cite{McIntyre2020, Lui2021}. It would be interesting to look into this correlation in our dataset as well; one could train a model to predict, on the basis of patient and prescriber data, whether benzodiazepines will be prescribed. If there are correlations between the gender of the prescriber and the prescription of benzodiazepines, we could raise a warning to let the practitioner know that the model thinks there might be a bias.

Finally, there are other medications for which experts suspect there could be gender biases in the prescriptions and administrations, such as antipsychotics and antidepressives. It would be beneficial to also study those administrations using a similar pipeline to the one developed here.

As a final note, \cite{Pfohl2021} warned against the use of blind applications of fairness frameworks in healthcare. Thus, the present study should be considered only as a demonstration of the importance of considering bias and mitigation in clinical psychiatry machine learning models. Further work is necessary to understand these biases on a deeper level, and what should be done about them.

\section{Conclusions}
\label{sec:5}
Given our results (Section~\ref{sec:3}) and discussion thereof (Section~\ref{sec:41}), we can conclude that a model trained to predict future administrations of benzodiazepines based on past data is biased by the patients' genders. Perhaps surprisingly, reweighing the data (a pre-processing step) seems to mitigate this bias quite significantly, without loss of performance. The in-processing method Prejudice Remover also mitigated this bias, but at a cost to performance.

This is the first fairness evaluation of a machine learning model trained on real clinical psychiatric data. Future researchers working with such models should consider computing fairness metrics and, when necessary, adopt mitigation strategies to ensure patient treatment is not biased with respect to protected attributes.



\vspace{6pt} 





\section{Notes}
The study was approved by the UMC ethics committee as part of PsyData, a team of data scientists and clinicians working at the psychiatry department of the UMC Utrecht.

The datasets generated for this study cannot be shared, to protect patient privacy and comply with institutional regulations.

The core content of this study is drawn from the Master in Business Informatics thesis of Jesse Kuiper~\cite{Kuiper2021}.

\bibliographystyle{plain}
\bibliography{biblio}

\end{document}